\documentclass[10pt,twocolumn,letterpaper]{article}

\usepackage{3dv}
\usepackage{times}
\usepackage{epsfig}
\usepackage{graphicx}
\usepackage{amsmath}
\usepackage{amssymb}
\usepackage{tabularx}
\usepackage{makecell}
\usepackage{amsfonts}
\usepackage{gensymb}

\threedvfinalcopy 

\def\threedvPaperID{****} 

\ifthreedvfinal\fi
\begin{document}

\title{Learning camera viewpoint using CNN to improve 3D body pose estimation}

\author{Mona Fathollahi Ghezelghieh, Rangachar Kasturi, Sudeep Sarkar\\
Department of Computer Science and Engineering\\
University of South Florida, Tampa\\
\tt (mona2, R1K, sarkar)@mail.usf.edu\\
}

\maketitle

\begin{abstract}
The objective of this work is to estimate 3D human pose from a single RGB image. Extracting image representations which incorporate both spatial relation of body parts and their relative depth plays an essential role in accurate 3D pose reconstruction. In this paper, for the first time, we show that camera viewpoint in combination to 2D joint locations significantly improves 3D pose accuracy without the explicit use of perspective geometry mathematical models. 

To this end, we train a deep Convolutional Neural Network (CNN) to learn categorical camera viewpoint. To make the network robust against clothing and body shape of the subject in the image, we utilized 3D computer rendering to synthesize additional training images.
We test our framework on the largest 3D pose estimation benchmark, Human3.6m, and achieve up to 20\% error reduction compared to the state-of-the-art approaches that do not use body part segmentation \footnote{To appear at the International Conference on 3D Vision (3DV), 2016.}.
\end{abstract}

\section{Introduction}

Estimating 3D human pose from an ordinary monocular image has been one of the active research areas for several years. It has a wide spectrum of applications in surveillance, human-computer interaction, gaming, activity recognition and virtual reality.
Researches in this area not only should overcome the challenges that exist in 2D pose estimation such as highly complex body articulation, clothing, lighting and occlusion, they should resolve the ambiguities that rise from the projection of 3D objects to the image plane. These limitations are mainly overcome by employing multiple synchronized cameras or exploiting motion information in consecutive frames \cite{compSeq}. However, there is still a great need to infer 3D pose from a single RGB image which is our focus in this paper.

\begin{figure}[t]
\begin{center}
   \includegraphics[width=1\linewidth]{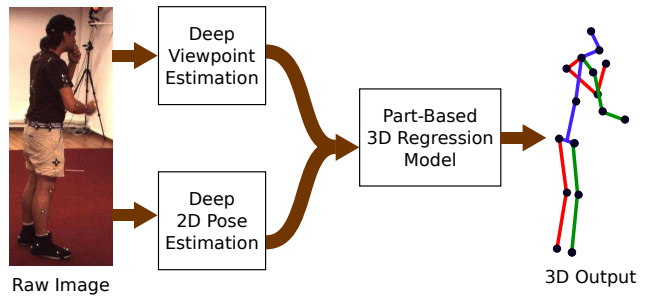}
\end{center}
   \caption{Estimated 2D pose and camera viewpoint are given to joint-set regression model to obtain 3D pose}
\label{fig:fw}
\end{figure}

Discriminative regression-based approaches such as \cite{twingp} \cite{catalin} \cite{3d-cnn} either make contributions in extracting new image features tailored to the task and/or build a new model to map these descriptors to 3D pose. While these approaches are effective at 3D pose reconstruction, there is still much room for improvement. The big challenge in this context, is to design or learn (in the case of CNN) rich features to encode both depth and spatial relation of body parts \cite{seq-tekin}.

We hypothesize that camera viewpoint in combination with 2D joint locations could resolve the problem in constructing human 3D pose from a monocular image. Camera viewpoint carries much  information on the relative depth of the person in the image. For example, if we could infer that the person orientation with respect to the camera is 45°, we could reason that the depth of his or her left shoulder is more than right shoulder. We will show that even categorical viewpoint angle (8 categories) has enough information that in combination with 2D joint locations significantly improves 3D pose estimation accuracy.

Recent progress in 2D pose estimation techniques \cite{ief}, both in terms of accuracy and speed, has removed the need to design or train a new 2D pose predictor for the target dataset.

The main challenge is how to learn a model to predict camera viewpoint. The predictor should be robust enough against fine-grained pose variations and only learn the coarse-grained body orientation. In addition, the prediction model should be invariant to the person body shape, background and lighting. Inspired by the great success of CNN in addressing these challenges \cite{scene} \cite{krizh}, we also adopted CNN for camera viewpoint estimation.

Additionally, to make the network invariant to the clothing texture, we propose to use synthetic dataset. To this end, we utilized 3D graphic software and CMU mocap dataset to create synthetic characters with different 3D poses and viewpoint annotations and illustrate its efficacy in making the network invariant to the clothing variations.

The novel contribution of this paper is therefore a principled approach to combine camera viewpoint and 2D joint locations to predict 3D body pose from monocular images. Furthermore, we demonstrate that training CNN with synthesized 3D human model makes it invariant to human body shape and clothing texture. 

We test our framework on the largest 3D pose estimation benchmark, Human3.6m \cite{h36m}, and achieve up to 20\% error reduction compared to the state-of-the-art approaches \cite{h36m} \cite{3d-cnn} \cite{max-margin} that do not use body part segmentation or a sequence of frames.

\section{Related work}
3D human pose estimation from a single RGB image is a challenging problem in computer vision. To estimate 3D pose accurately, it is critical to use expressive image representation. One way to categorize previous methods is based on whether they directly extract image features or utilize an existing method to estimate 2D joints' landmarks. 

There is a large literature belonging to the first group. For example, in \cite{catalin} first body parts are segmented and then are described by second-order label-sensitive pooling \cite{pooling};  the approach in \cite{twingp} represents the image with HOG features; and LAB and HOG features are used in \cite{sweep}. 
Convolutional neural network has also been exploited to learn image features and regression model simultaneously; for example two neural networks are trained in \cite{max-margin}  to learn image features and 3D pose embedding which are later used to learn a score network that can assign high score to correct image-pose pairs and low scores to other pairs.  Li et al  \cite{3d-cnn}  proposed a CNN multi-task framework that jointly learns pose regression and body part detectors.

The challenge in approaches is that the image feature should be rich enough to represent both pairwise relationships between joints in 2D space and their relative depth information. 
 
Techniques which fit in the second category utilize an already existing method to estimate 2D joint locations. The reconstructed 3D poses should be disambiguated to account for missing depth information. In the following we review some of the representative approaches in this group.

In \cite{sheik}, 3D human pose is represented as a sparse embedding in an overcomplete dictionary. The authors proposed a matching pursuit algorithm to sequentially select E basis poses that minimize the reprojection error and refine the projective camera parameters.  Fan et al  \cite{usc} extended this work by hierarchically clustering the 3D dictionary into subspaces with similar poses. To reconstruct the 3D pose from a 2D projection, the selected pose bases are drawn from a small number of subspaces that are close to each other.  

Yasin et al in \cite{dual-source} combined two different datasets to generate many 3D-2D pairs as training examples. During inference, estimated 2D pose is used to retrieve the normalized nearest 3D poses. The final 3D pose is then estimated by minimizing the projection error under the constraint that the estimated 3D should be close to the retrieved poses.
 Akhter et al \cite{black} proposed a new framework to estimate 3D pose from ground truth 2D pose. To resolve the ambiguity, they first learn the pose-dependent joint angle limits by collecting a new mocap dataset which includes an extensive variety of stretching poses.  
Radwan et al in \cite{occlusion}, imposed kinematic constraint through projecting a 3D model onto the input image and pruning the parts which are incompatible with the anthropomorphism. To reduce the depth ambiguity, several 3D poses were generated by regressing the initial view to multiple oriented views. Estimated orientation from 2D body part detector is used to choose the final 3D pose. Simo-Serra et al in \cite{2d3d} proposed a Bayesian framework to jointly estimate both the 3D and 2D poses. The set of 3D pose hypotheses are generated using 3D generative kinematic model, which are weighted by a discriminative part model.

Our proposed approach has the advantages of both categories; we directly use estimated 2D joint locations to account for spatial relation of body parts, and learn camera viewpoint to incorporate depth information.

\begin{figure*}[t]
\begin{center}
   \includegraphics[width=0.6\linewidth]{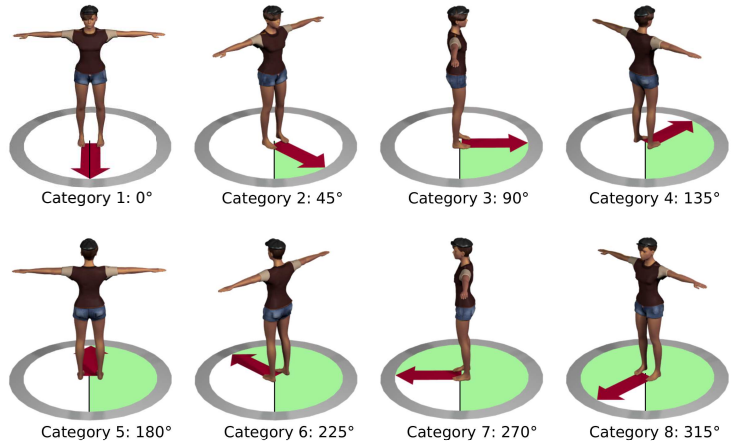}
\end{center}
   \caption{Camera viewpoint categorizaion. The angles displayed in the figure are obtained by discretizing the yaw angle of human subject}
\label{fig:angles}
\end{figure*}

\section{Method}
The goal of our model is to extract 3D human pose with respect to the camera in a single RGB image. Figure \ref{fig:fw} outlines our approach which can be split into three major parts: 2D joint localization, viewpoint estimation and regression model.

\subsection{Viewpoint estimation}\label{sec:viewpoint-estimation}
Human viewpoint estimation with respect to the camera has many applications by itself; for example, in \cite{parse} and \cite{HCI}, person orientation is defined as a human attribute for a robotics/automotive application scenarios. 
In addition, it carries much information on the depth of body parts which makes it a good candidate in resolving depth ambiguity in estimating 3D pose. For example, if we know that the orientation is 90 degree we can conclude that depth of left hand is more than the depth of the right hand.  
In our proposed framework, we discretize the viewpoint angle into eight bins $(0\degree , 45\degree , \dots, 315\degree)$ as shown in Figure \ref{fig:angles} and define viewpoint estimation as a classification problem.

Viewpoint estimation is a challenging task due to the wide variety of clothing, body size, background and poses in the same viewpoint. In the past, some researches have approached this problem with designing HOG features, which yields decent performance for simple scenarios such as a walking person. However, these handcrafted features are not expressive enough in our application where there is no restriction on the human activity. For this reason, we have considered CNN framework which has shown a good performance in learning hierarchical and contextual features in other computer vision tasks such as classification \cite{krizh}, scene labeling \cite{scene} and speech processing \cite{dahl2012context}.

\textbf{Method}: Assuming that the person is in the center of the image, we aim to train a CNN to infer the orientation of the person regardless of other pose variations, clothing and background. This is a difficult task and needs a large dataset that incorporates all these variations. On the other hand, due to the large number of parameters, CNNs are prone to overfit on smaller datasets. This can be alleviated to some extent by pre-training the weights on a large-scale task, followed by training on the target task (fine-tuning).
We adopted Alexnet architecture \cite{krizh}, and initialize the weights from a model pre-trained on the Imagenet \cite{imagnet} classification task. Only class-dependent fully connected layers are trained from the scratch. 

In our 3D pose estimation framework, the predicted category is then mapped to the viewpoint angle and is concatenated to the image features. The naive approach is to directly append viewpoint to the 2D features, but this could cause distance ambiguities. For example, let us assume that the first image is frontal, $\theta=0\degree$, the orientation of the second image is $\theta=315\degree$ and third image is backward $\theta=180\degree$. In this case, the first image is more similar to the second image than the third image in terms of viewpoint angle (see Figure \ref{fig:angles}), while the distance between the first and second image in feature space is $\Delta_{12}=315$ and the distance of first and third is $\Delta_{13}=180$. It means $\Delta_{12}< \Delta_{13}$ which is not valid.
To resolve this problem, we map the viewpoint angle to $(\sin \theta, \cos \theta)$ vector. Therefore $\Delta_{12}= 0.76$, $\Delta_{13}=1.41$, which yields $\Delta_{12}< \Delta_{13}$. 
This vector is further scaled by a fixed coefficient $M$ to account for the influence of viewpoint in 2D feature representation. In our experiments, $M$ is chosen to be 100 to make the viewpoint features comparable to 2D features.

There are two main scenarios to train and evaluate our deep viewpoint network:
\begin{enumerate}
\item Within subjects: Train and test on the same set of subjects. This is an easy scenario where the training set include some images from the test subjects.
\item Across subjects: Test subjects are different from the subjects present in the training set. This scenario is much harder than the first one, because the network should learn the viewpoint and being invariant to subjects clothing texture and body shape.
\end{enumerate}

\begin{figure*}
\begin{center}
   \includegraphics[width=1\linewidth]{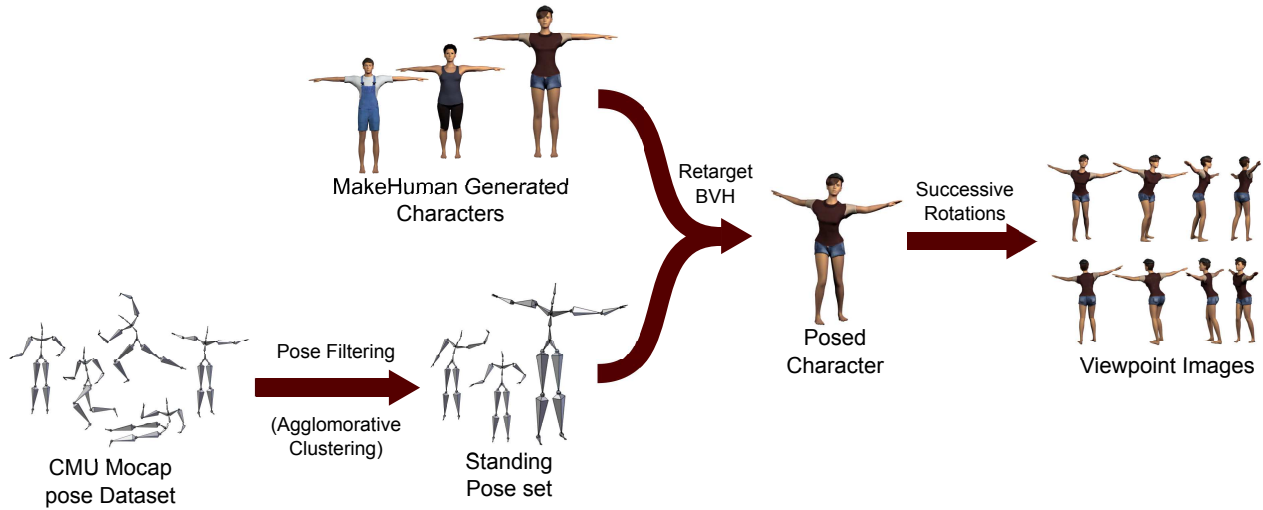}
\end{center}

   \caption{Training samples synthesis with different camera viewpoints}
\label{fig:render}
\end{figure*}

\subsection{Training Data Rendering}\label{sec:train-data-rend}

To train the network to be invariant to the human appearance and only learn the camera viewpoint, many  training samples with different clothing texture or body shape should be collected and be annotated with camera viewpoint which is an expensive task. 

Similar to the approaches in  \cite{3Dchair}, \cite{chen2016synthesizing}, \cite{genChair} and \cite{renderedCNN}, we render 3D human characters with different clothing and skeleton shapes and in various 3D poses.   
To this end, CMU-mocap dataset\footnote{The CMU data was obtained from                       http://mocap.cs.cmu.edu} in BHV (Biovision Hierarchy) format is used to generate different poses. In this paper, we only consider 3D pose estimation for upright poses.  In the future we plan to extend our approach to more complex activities such as sitting down or lying down. For this reason, to select only upright poses of CMU-mocap dataset we perform agglomerative clustering on the BVH angular rotation vectors. The largest cluster which has 760 standing poses is used as 3D skeleton of the rendered characters. 

We have also used MakeHuman, an open source 3D computer graphic software to generate characters with different attributes (gender, height, etc.) and clothing. The 3D morphing feature of MakeHuman \cite{ref} facilitates human character creation with variant attributes and clothing which would be a very time-consuming task otherwise.

Each 3D pose is then applied to all characters using Retarget feature of MakeWalk add-on in Blender software. Furthermore, different viewpoint images are obtained by human character rotations in 45° angular steps. Figure \ref{fig:render} illustrates our framework to generate synthesized training examples to train our deep viewpoint CNN.

\subsection{2D pose estimation}
Viewpoint estimation network provides a coarse estimation of 3D pose. Therefore, to accurately estimate the 3D coordinates of human joints with respect to the camera, accurate information of body parts in the image is required. We use Iterative Error Feedback (IEF) \cite{ief} to estimate the x,y location of body joints in the image.

IEF is a CNN based approach with a feedback structure that learns the corrections to the initial predefined pose towards the true 2D pose. The network input is RGB image augmented with $Nj$ image planes where each image is a heat-map of one of the predicted body joints ($Nj$ is the number of joints). In the first iteration, these augmented planes are initialized by pre-defined joints and in the following iterations the network learns what corrections should be made to these initial joint locations and updates them.

\begin{figure}[b]
\begin{center}
   \includegraphics[width=1\linewidth]{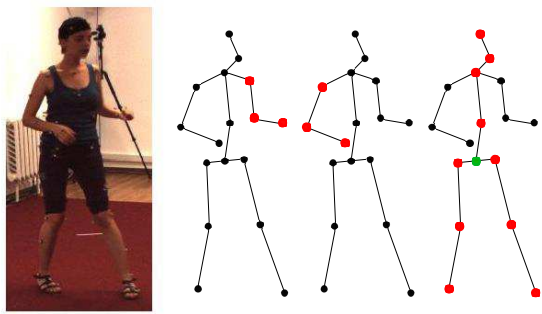}
\end{center}
   \caption{Joint sets used in our regression model. The reference node is shown with green color.}
\label{fig:joints}
\end{figure}

\subsection{Part-based 3D regression model}
Our objective in this section is to learn a mapping from 2D features (joint locations in RGB image and camera viewpoint) to the corresponding 3D pose (target space).  While this approach is similar to many other computer vision problems such as object recognition and scene classification, the main difference is the strong correlation among target variables. For example, when the person performs a particular action, his or her joints movement are highly correlated. 

To account for the dependency in both input and target space, we adopted the regression method proposed in \cite{twingp} which enforces that the distributions of similar inputs (2D features) and similar outputs to be close. This was achieved by minimizing the Kullback-Leibler (KL) divergence between the distribution of outputs and input features.

Following the notations in \cite{twingp}, the input features are denoted by $\mathbf{r}$ and the corresponding 3D pose by $\mathbf{x}$. Training inputs and outputs are represented by $R=(r_1,r_2,\dots,r_N)$ and $X=(x_1,x_2,\dots,x_N)$ respectively, where each one was modeled by a Gaussian process \cite{GP}. Therefore the joint distribution of test input, $\mathbf{r}$, and training inputs, $R$, is given by,
\begin{align}
  \mathcal{N}_R \left( 0, \left[ \begin{array}{cc}K_R & K_R^r \\ (K_R^r)^T & K_R(\mathbf{r},\mathbf{r}) \end{array} \right] \right)
\end{align}

where $K_R$ is the $N \times N$ covariance matrix of training features and $K_R^r$ is a covariance function of test input with training inputs ($N \times 1$ vector). By employing RBF kernel in calculating covariance function its $(i,j)$-th element is represented by,
\begin{align}
  K_R(\mathbf{r}_i, \mathbf{r}_j) = \exp \left( - \gamma_r \| \mathbf{r}_i - \mathbf{r}_j \|^2 \right) + \lambda_r \delta_{ij}.
\end{align}

where $\gamma_r$ is the kernel width parameter, $\lambda_r$ is the variance of noise and $\delta_{ij}$ is the Kronecker delta function.

Similarly the output 3D pose distribution can be modeled by Gaussian Process, where the covariance matrix of training 3D poses is represented by $K_X$ and $K_X^x$ is a $N \times 1$ column vector defined as
\begin{align}
(K_X^x)_i = K_X(\mathbf{x}_i, \mathbf{x})
\end{align}
where $\mathbf{x}_i$ is the $i$-th 3D pose in the training set. 

Following the derivations in \cite{twingp}, the  Kullback-Leibler divergence measure is given by
\begin{align}
  L(\mathbf{x}) &= D_{KL} (\mathcal{N}_X \| \mathcal{N}_R) \nonumber \\ 
&= K_X(\mathbf{x}, \mathbf{x}) - 2(K_X^x) ^ \intercal K_R^{-1} K_R^r \nonumber \\ 
&- \left[K_R(\mathbf{r}, \mathbf{r}) - (K_R^r) ^ \intercal K_R^{-1} K_R^r  \right] \nonumber \\ 
&\times \log\ \left[K_X(\mathbf{x}, \mathbf{x}) - (K_X^x) ^ \intercal K_X^{-1} K_X^x  \right] \nonumber \\ 
\end{align}

Therefore, estimated 3D pose, $\mathbf{x}^*$, is obtained by minimizing this divergence measure  \cite{twingp}
\begin{align}
  \mathbf{x}^* = \arg \min_{\mathbf{x}} [L(\mathbf{x}) \equiv D_{KL} (\mathcal{N}_X \| \mathcal{N}_R)]
\end{align}

The performance of this approach depends on the availability of similar training feature to the test feature. Of course collecting more training samples is one way to tackle this challenge, but since humans have much articulation capabilities it is almost impossible to capture all pose variations in the training set. For example, the training set might include examples of walking pose with hands in pockets, while the test image is a walking person who is waving.
In this paper, for the first time, we present the idea of joint set regression model, that tries to alleviate this problem to some extent. To this end we divide human joints into three classes: right-hand, left-hand, head-torso, as shown in Figure \ref{fig:joints}. Therefore, three different regression models are learned corresponding to the right-hand, left hand and torso and lower body. In the experiment section, we will show this will yield reduction of 3D pose reconstruction error while increasing the complexity linearly (by a factor of 3). The joints in the right and left legs could also be separated, however our experiments shows marginal improvement. 

\begin{table*}[t]
\centering
\caption{Error of CNN based camera viewpoint estimation.}
\label{tab:viewpointerr}
\begin{tabular}{lcc}
\hline
& \multicolumn{2}{c}{Scenario}  \\\cline{2-3}
Training dataset                  & \multicolumn{1}{l}{Within subjects} & \multicolumn{1}{l}{Across subjects} \\ \hline
H80K dataset                     & 8.5\%                                & 34\%                                 \\ \hline
Centered H80K dataset            & 5\%                                  & 30\%                                 \\ \hline
H80K dataset + Synthetic dataset & 3.7\%                                & 20\%                                 \\ \hline
\end{tabular}
\end{table*}

\section{Experimental results}
Experiments were performed to investigate the feasibility of the proposed 3D pose estimation framework. In the first subsection, we evaluate the accuracy of our proposed deep camera viewpoint estimation independent of 3D pose estimation. In the second subsection, the efficacy of each step is investigated. Finally,  3D pose estimation accuracy is studied and the results are compared with the other state-of-the-art approaches.

\textbf{Dataset}: Earlier datasets on human 3D pose estimation such as “Human Eva” \cite{eva} is still commonly used for evaluations in the literature. However, the limited size of the training set, relatively simple test scenarios, non-challenging clothing texture of the subjects and lighting make this dataset unsuitable for training a CNN based model. Therefore, we have used Human3.6m dataset  \cite{h36m}, which includes video recordings of 11 different subjects performing motion scenarios based on typical human activities such as Direction, Discussion, Eating, Sitting on chair, etc. Since our objective is to estimate 3D pose from a single RGB image, we follow the procedure used in  \cite{catalin} and use downsampled subset of this dataset which is called H80k. Furthermore, in this paper only activities that consist mainly upright poses are considered i.e. Direction, Discussion, Greeting, Walking and Walking together.

In this dataset, 3D body poses are represented by skeletons with 17 joints defined in the coordinate system of the camera that captured the images. We use the relative coordinates with respect to the pelvis joint to be consistent with other works mentioned in this section.

\subsection{Accuracy of deep camera viewpoint estimation}
Even though there are several datasets with viewpoint annotations in depth domain, to the best of our knowledge there is no publicly available dataset in RGB domain. Therefore, to generate training set we have annotated H80K dataset with categorical viewpoint labels. For this purpose, yaw angle is calculated using 3D coordinates of right and left shoulders for each image, then the calculated yaw angles are discretized into eight orientation bins based on the pre-defined intervals (Figure \ref{fig:angles}). For example, if the yaw angle is between -5 and +5 it belongs to the class 1 or $\theta=0\degree$ orientation. 
In addition, to make our training images cleaner and more specific to the task, we first filter all images that are not upright pose. For this purpose, we employ hierarchical clustering based on the 3D coordinates of feet and torso to cluster data into three different groups. The largest cluster is selected as the training set in this paper.

MatconvNet CNN library \cite{matconv} is utilized to train and test our Convolutional Neural Network. 
Two pre-trained deep networks, VGG-f and VGG-m \cite{devil}, are fine-tuned on our dataset. These networks have already been pre-trained on ImageNet ILSVRC-2012 challenge dataset \cite{imagnet}. While VGG-M is more accurate on object classification, it consistently had worse performance on our viewpoint estimation task. This might be due to the fact that the convolutional filters in VGG-M are more tuned to recognize objects and therefore are more sensitive to clothing texture for example. All the experiments have been done with batch size of 100 and learning rate of 10e-3. 

 The performance of our deep viewpoint estimation is shown in Table \ref{tab:viewpointerr}. Two networks are trained and tested corresponding to two different scenarios (section \ref{sec:viewpoint-estimation}): “Within Subject scenario” and “Across Subjects scenario”, where the subjects in training set are different from the ones in the test set.

 Each scenario is evaluated in three different setups. In the first one, we utilize the upright poses of H80K dataset. The training data is reduced to 8800 samples due to viewpoint discretization explained in section \ref{sec:viewpoint-estimation}. In the next setup, the training and test images are centralized based on a fixed point in the torso. This slightly improves the accuracy in both “Within Subjects” and “Across Subjects” scenarios.  In the third experiments we use our synthesized dataset in combination with H80k dataset for training our CNN. Ten different 3D human characters with different body shape and clothing in upright poses are rendered and are annotated with viewpoint category (section \ref{sec:train-data-rend}). This yields significant improvement in “Across Subjects” experiment.

\subsection{3D pose estimation results}
The proposed predictive model has three components: camera viewpoint estimation, 2D pose estimation, and joint-set regression model. In this section, implementation details of each component are explained and their influence on the 3D pose reconstruction error are studied. 

\textbf{Error measure:} Similar to other papers reporting on 3D pose estimation benchmark, we calculate MPJPE (Mean Per Joint Position Error) metric. For each image, this metric is given by 
\begin{align}
  E_{\text{MPJPE}} = \frac{1}{N_S}\sum_{i=1}^{N_s} \| \mathbf{m}_{\text{est}} (i) - \mathbf{m}_{\text{gt}} (i) \|_2
\end{align}
Where $N_S$ is the number of joints in the skeleton, $\mathbf{m}_\text{gt} (i)$ is the 3D coordinate of $i$th joint and $\mathbf{m}_\text{est} (i)$ is the estimated coordinate.

\begin{figure}[b]
\begin{center}
   \includegraphics[width=0.95\linewidth]{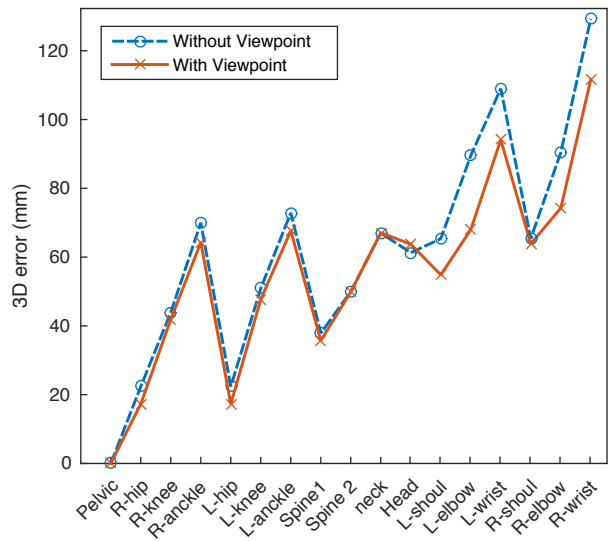}
\end{center}
   \caption{3D error of each body joint.}
\label{fig:lineplot}
\end{figure}

\subsubsection{Influence of viewpoint estimation }
Figure \ref{fig:lineplot}  depicts the results of mean 3D pose estimation per body joint for the “Direction activity”, with and without viewpoint incorporation. We observe a significant improvement in the right and left hand joints but only modest improvement in the legs and torso.   
This can be attributed to the higher degree of freedom in hands compared to legs and torso, which leads to a higher chance of ambiguity in inferring 3D from 2D coordinate. This has been, to some extent, addressed by our camera viewpoint estimation.

\subsubsection{Influence of non-perfect 2D pose}
\begin{figure}[t]
\begin{center}
   \includegraphics[width=1\linewidth]{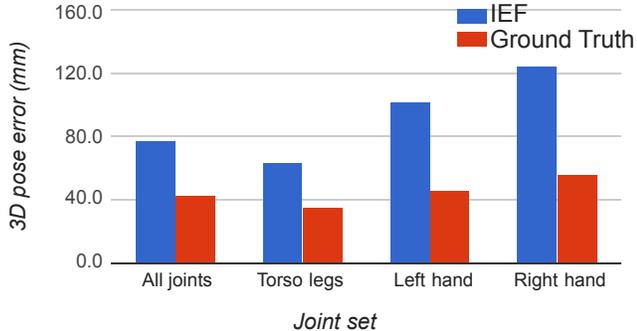}
\end{center}
   \caption{Effect of non-perfect 2D pose on 3D pose reconstruction error.}
\label{fig:errors}
\end{figure}

In this subsection, we investigate the effect of non-perfect 2D pose estimation. To make the experiments isolated from our viewpoint estimation, ground truth viewpoint is used in this experiment. Figure \ref{fig:errors} illustrates the mean 3D pose estimation error for each joint set. We observe that the performance drops even more for the right and left hand joints. Part of this performance drop could be due to the frequent occlusion of the hands.

\subsubsection{Influence of joint set regression}
In this subsection, we show that our joint set regression is more effective compared to the approach that estimates all joints with one regression model. Similar to the previous experiment, in this part ground truth viewpoint is used in estimating the 3D pose, to isolate the effect of regression from viewpoint estimation. Table \ref{tab:regr} shows our results on validation set subjects. Our ‘joint-set’ regression model improves the accuracy in 3D pose estimation for both subjects.

\begin{table}[b]
\centering
\caption{Effect of separate model for each joint set on 3D pose estimation error (mm).}
\label{tab:regr}
\begin{tabular}{llll}
\hline
Regression Model & All-joint & Joint-set & Improvement \\ \hline
Subject 5        & 74.1      & 70.45     & 5.2\%       \\ \hline
Subject 6        & 105.7     & 99.89     & 5.8\%       \\ \hline
\end{tabular}
\end{table}

\subsection{Comparison with state-of-the-art methods}
Finally, we evaluate our automatic 3D pose estimation framework with a baseline method and a few state-of-the art methods in two different experimental setups. The baseline method \cite{h36m} describes each image by Fourier approximation of its HOG features followed by a Regressor based on Kernel Dependency Estimation, where both input features and output 3D poses are transferred into high-dimensional Hilbert spaces; then a linear function is learned to model the dependency between them. In the following tables, this method is referred to as “eχ2 - HOG+KDE”.

In \textbf{Subject Specific} scenario, each subject is considered separately, i.e., for each subject the test set includes all images of activity x and training set is the rest of activities of the same subject. Therefore, we use our “within subject” viewpoint estimation network. The focus of this experiment is on the pose variations; the body shape and clothing are not changing in training and test sets.

\begin{figure}[!b]
\begin{center}
   \includegraphics[width=0.7\linewidth]{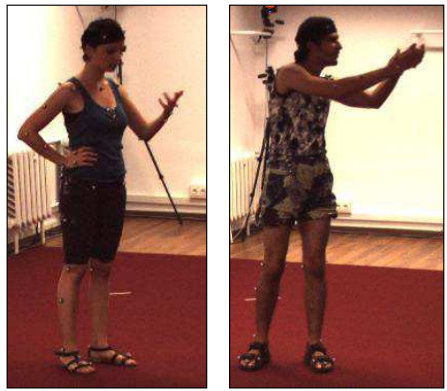}
\end{center}
   \caption{Examples of validation set images S5 and S6 from left to right.}
\label{fig:subjects}
\end{figure}

This experiment was performed for subjects “S5” and “S6” in the validation set shown in Figure \ref{fig:subjects}. The results are summarized in Table \ref{tab:ssm} where our approach outperforms the baseline method. An interesting observation is that the error for subject “S6” is slightly higher than “S5”. We have observed that both the 2D pose estimation and camera viewpoint prediction errors were higher for this subject. We believe that this is mainly due his clothing texture.

\begin{table}[t]
\centering
\caption{Mean 3D pose estimation error (mm) in Subject Specific Model}
\label{tab:ssm}
\begin{tabularx}{\columnwidth}{lll}
\hline
Method                        & S5 & S6 \\ \hline
eχ2 - HOG+KDE \cite{h36m}            & 96.35      & 113.8    \\ \hline
DeepViewPnt(ours) & 71.66      & 99.89    \\ \hline
\end{tabularx}
\end{table}

\begin{table*}[!t]
\centering
\caption{Mean 3D pose estimation error (mm)  in Activity Specific Model (Only upright activities are evaluated).}
\label{tab:asm}
\begin{tabular}{lccccccc}
\hline
Method                  & \multicolumn{1}{l}{Single Image}   & \multicolumn{1}{l}{Direction} & \multicolumn{1}{l}{Discussion} & \multicolumn{1}{l}{Eating} & \multicolumn{1}{l}{Greeting} & \multicolumn{1}{l}{Walking} & \multicolumn{1}{l}{Walking together} \\ \hline
RSTV-DN \cite{compSeq}          & X        & 102.41                         & 147.72                          & 88.83                       & 125.28                        & 55.7                        & 65.76                                \\ \Xhline{3\arrayrulewidth}
eχ2 - HOG+KDE\cite{h36m} & \checkmark              & 115.79                         & 113.27                          & 99.52                       & 128.80                        & 131.15                       & 146.14                                \\ \hline
DconvMP-HML \cite{3d-cnn}       & \checkmark          & -                              & 148.79                          & 104.01                      & 127.17                        & 77.60                        & -                                     \\ \hline
StructNet-Avg \cite{max-margin}      & \checkmark          & -                              & 92.97                           & \textbf{76.70}              & 98.16                         & 99.40                        & 109.30                                \\ \hline
\textbf{DeepViewPnt (ours)} & \checkmark       & \textbf{80.30}                 & \textbf{80.39}                  & 78.13                       & \textbf{89.72}                & \textbf{95.07}               & \textbf{82.22}                        \\ \hline
\end{tabular}
\end{table*}

In \textbf{Activity Specific} Scenario each activity is considered separately. In this scenario, the regression model is trained on activity “x” of all training subjects and is tested on the activity “x” of the test subjects.  This is a very challenging scenario because the subjects in the test set are different from training subjects both in terms of body shape and clothing texture. Therefore, we employ our “Across Subjects” network to first estimate camera viewpoint which in itself has high error rate compared to our “Within Subjects“ network in previous section. 
Our comparison with the state-of-the-arts for all upright activities in H3.6M dataset is summarized in Table \ref{tab:asm}. All these methods are based on a single image, except RSTV-DN \cite{compSeq} which exploits consecutive frames in 3D pose estimation. Our approach outperforms or is comparable to the single image based approaches.

Please note that in all of these methods, entire 3.6 million images of H3.6M dataset are used for training, while we used H80k dataset for training which is a downsampled version of this dataset. 

\section{Conclusions and Future work}
 
We have proposed a CNN based approach to estimate the categorical camera viewpoint, which by itself is useful to infer the coarse-grained human 3D pose. We have illustrated, for the first time, that training a CNN using additional synthetic human models with various clothing textures and skeleton shapes improves the viewpoint prediction accuracy when the character in the test image is not present in the training set.

The estimated camera viewpoint provides strong clue such that its combination with the state-of-the-art 2D pose estimator significantly improves 3D pose reconstruction accuracy in monocular images. We achieved state-of-the-art performance on the largest 3D pose estimation benchmark.

Future work should consider 3D pose estimation in non-laboratory environments; for example, scenarios where multiple people are present in the image, the person is interacting with an object, and the dataset includes more diverse background.
In addition, this framework could be extended to estimate 3D pose in non-upright activities such as sitting on the chair or laying on the ground.
Finally, camera viewpoint estimation could be more robust and accurate if several consecutive frames are considered.
Estimating the number of frames that are required to make an accurate decision is another possible future direction of this work.

{\small
\bibliographystyle{ieee}
\bibliography{egbib}
}

\end{document}